# How to predict on-road air pollution based on street view images and machine learning: a quantitative analysis of the optimal strategy

Hui Zhong [a,b,1], Di Chen [a,1], Pengqin Wang [c], Wenrui Wang[a], Shaojie Shen[d], Yonghong Liu[f,*], Meixin Zhu [a,b,e,*]

[a] *Intelligent Transportation Thrust, Systems Hub, The Hong Kong University of Science and Technology (Guangzhou), Guangzhou, China;*

[b] *Department of Civil and Environmental Engineering, The Hong Kong University of Science and Technology, Hong Kong, China;*

[c] *Division of Emerging Interdisciplinary Areas (EMIA), Interdisciplinary Programs Office, The Hong Kong University of Science and Technology, Hong Kong, China;*

[d] *Department of Electronic and Computer Engineering, the Hong Kong University of Science and Technology, Hong Kong SAR, China*

[e] *Guangdong Provincial Key Lab of Integrated Communication, Sensing and Computation for Ubiquitous Internet of Things, Guangzhou, China;*

[f] *School of Intelligent Systems Engineering, Sun Yat-sen University, Guangzhou, 510006, China;*

**Abstract:** On-road air pollution exhibits substantial variability over short distances due to emission sources, dilution, and physicochemical processes. Integrating mobile monitoring data with street view images (SVIs) holds promise for predicting local air pollution. However, algorithms, sampling strategies, and image quality introduce extra errors due to a lack of reliable references that quantify their effects. To bridge this gap, we employed 314 taxis to monitor NO, $NO_2$, $PM_{2.5}$ and $PM_{10}$ dynamically and sampled corresponding SVIs, aiming to develop a reliable strategy. We extracted SVI features from ~ 382,000 streetscape images, which were collected at various angles (0°, 90°, 180°, 270°) and ranges (buffers with radii of 100m, 200m, 300m, 400m, 500m). Also, three machine learning algorithms alongside the linear land-used regression (LUR) model were experimented with to explore the influences of different algorithms. Four typical image quality issues were identified and discussed. Generally, machine learning methods outperform linear LUR for estimating the four pollutants, with the ranking: random forest > XGBoost > neural network > LUR. Compared to single-angle sampling, the averaging strategy is an effective method to avoid bias of insufficient feature capture. Therefore, the optimal sampling strategy is to obtain SVIs at a 100m radius buffer and extract features using the averaging strategy. This approach achieved estimation results for each aggregation location with absolute errors almost less than 2.5 μg/m³ or ppb. Overexposure, blur, and underexposure led to image misjudgments and incorrect identifications, causing an overestimation of road features and underestimation of human-activity features, contributing to inaccurate NO, $NO_2$, $PM_{2.5}$ and $PM_{10}$ estimation. These findings provide a better understanding and valuable support for developing image-based air quality models and other SVI-related research.

[*]Corresponding author
*E-mail address:* meixin@ust.hk. (Meixin Zhu)
*E-mail address:* liuyh3@mail.sysu.edu.cn. (Yonghong Liu)
[1]Equal contribution

## 1. Introduction

Air pollution presents a substantial global health risk, leading to an estimated 7 million premature deaths annually (WHO, 2022). Studies indicate that on-road exposure levels typically exceed those at home or in urban background locations by 2-5 times (EPA, 2011; Liu et al., 2021; Zhong et al., 2023). This underscores the hazardous levels of pollution in road environments, emphasizing the need for monitoring intra-urban variations in on-road air pollution to accurately assess human exposure to different pollutants and the resulting health effects.

Fixed ambient monitoring stations are frequently employed in urban areas to measure regulatory pollutants, including carbon monoxide (CO), nitrogen oxides (NO and $NO_2$), and particulate matter ($PM_{2.5}$ and $PM_{10}$) for legal compliance (Castell et al., 2017) and scientific analysis (Rabl and de Nazelle, 2012). However, such stations are sparse or nonexistent in many developing regions (Apte et al., 2018). Even in the U.S., there are only 2-5 monitors per million inhabitants and 1,000 km² (Lim et al., 2012). Furthermore, pollution level can vary significantly over short distances (0.01-1 km) due to emission source distribution, dilution, and physicochemical processes (Karner, 2010). In response to these limitations, mobile monitoring has gained attention for capturing urban air pollution using vehicles equipped with monitoring devices (Kerckhoffs et al., 2022). For instance, Apte et al. (2017) implemented a mobile measurement approach that provided urban air pollution patterns with a spatial precision of 4−5 orders of magnitude greater than what fixed-site ambient monitoring can achieve. Similarly, Wu et al. (2020) used a taxi-based mobile atmospheric monitoring system to map air pollution levels in Cangzhou, China. Despite their effectiveness, these studies are subjected to specific cities or regions due to high costs.

The potential of street view images (SVIs) for urban analysis has been increasingly recognized, especially in studying urban structures and enhancing land use models (Hong et al., 2020; Kim et al, 2024). Numerous studies have employed these images via segmentation to create land-use regression (LUR) models for mapping road pollution cost-effectively (Ganji et al., 2020; Qi and Hankey, 2021; Li et al., 2022; Xu et al., 2022). These results demonstrate that using SVIs for modeling can be highly effective in developing comprehensive large-scale air quality models. Typically, this research follows a two-step process: initially, sampling SVIs using specific algorithms such as equidistant sampling; subsequently, utilizing these images to develop algorithms for urban prediction or sensing tasks.

Nevertheless, these studies often focus on the influence of segmentation variables from SVIs on model predictions, while neglecting the impact of various sampling strategies and prediction methods. For example, low-quality images captured in strong daylight or obstructed by vehicles contain minimal street information, directly affecting prediction results. Additionally, algorithms, sampling angles and ranges introduce extra errors without a reliable reference for quantifying their effects (Zhang et al., 2023). There is a notable lack of systematic study qualifying the effects of various strategies on errors in local air pollution prediction. To bridge this gap, this study employed 314 taxis to monitor air pollution dynamically and sampled corresponding SVIs, aiming to develop a reliable optimal strategy. The results can provide valuable insights and supports for urban planners and policymakers to develop empirical air pollution models and SVI-based research.

## 2. Materials and Methods

This study focuses on a major metropolis, Guangzhou, whose central urban area encompasses the Yuexiu, Haizhu, Tianhe, and Liwan districts. They serve as the city's primary political, economic, and cultural hub (Lin et al., 2011; Guangzhou Statistics Bureau, 2023). Detailed information regarding data collection and research methods is provided below.

### 2.1 On-Road Monitoring of Air Pollution

To collect pollutant data within the study area, we equipped 314 taxis with monitoring devices operating in Guangzhou from February 1, 2023, to March 31, 2023, yielding 15,534,069 entries and covering 1,440 hours and 562.6 km. Each participating taxi was fitted with sensors to measure four common pollutants: $NO_2$, $NO$, $PM_{2.5}$, and $PM_{10}$. Additionally, GPS units were installed to simultaneously collect geographic coordinates and trajectories. These above measurements were transmitted at interval of 15 s, which results in a comprehensive air pollution dataset with numerous repeated observations, ensuring the spatial and temporal balance. After cleaning and processing, 4,948,120 valid records were retained. In order to organize data conveniently, we applied spatial aggregation using 200×200 m grids along the road network, calculating the median concentrations of observations for each aggregation location (n = 4617). Referring to the calibration method from Hankey et al. (2019), we adjusted the mobile monitoring concentrations using the multiplicative background method to correct for potential biases due to daily diurnal variations in urban background concentrations. Firstly, the 15-s mobile data were aggregated as hour concentration for each aggregation location. Then, adjustment factors were calculated based on the ratio of the daily concentration to the hourly concentration at regulatory fixed-site monitors. Subsequently, these factors were applied to all mobile monitoring observations.

*2.2 SVIs Acquisition and Feature Extraction*

We utilized the Baidu API (https://map.baidu.com/) to acquire street view images (SVIs) for each aggregation location, capturing relevant on-road features. The API obtained SVIs at designated locations with a resolution of 512×1,024 pixels by inputting specific coordinates (longitude and latitude) and specifying sampling angles. To ensure sufficient SVI sampling, we applied a circular buffer sampling strategy (Apte et al., 2017). For each aggregation location, the centroid was considered the center of the corresponding buffer with different radius. We collected SVIs along road network as small as possible at interval of 25 m. Previous studies have indicated that the sampling angle and range (buffer size) for SVIs affect model performance (Hankey and Marshall, 2015; Zhang et al., 2023). To assess their influence, SVIs were captured from four directions: 0°, 90°, 180°, and 270° relative to the road orientation for each aggregation location within buffer radii ranging from 100m to 500m at 100m intervals. A total of 382,755 SVIs were collected in the study area. Figure 1 illustrates an example of the obtained SVIs, detailing the sampling strategy and the feature extraction process.

In addition, SVI quality can significantly impact model performance under consistent conditions (Zhang et al., 2023). Some images suffer from blurriness, overexposure, or color distortion due to lighting and environmental factors. To detect low-quality SVIs, we used the OpenCV image processing method (Garcia et al., 2015). The Laplace operator was employed to obtain the variance of the Laplace convolution result, helping to identify image blur. Generally, higher variance indicates more edge changes in the image, implying clearer SVIs. To recognize low-quality SVIs caused by lighting, images were converted to the HSV color space. By estimating the brightness and darkness channels, we detected overexposed and underexposed pixels, defining images with more than 50% of such pixels as low-quality. Additionally, we calculated the color histogram of each image channel to identify color distortion. These methods effectively classified low-quality SVIs independent of rotation and shift operations.

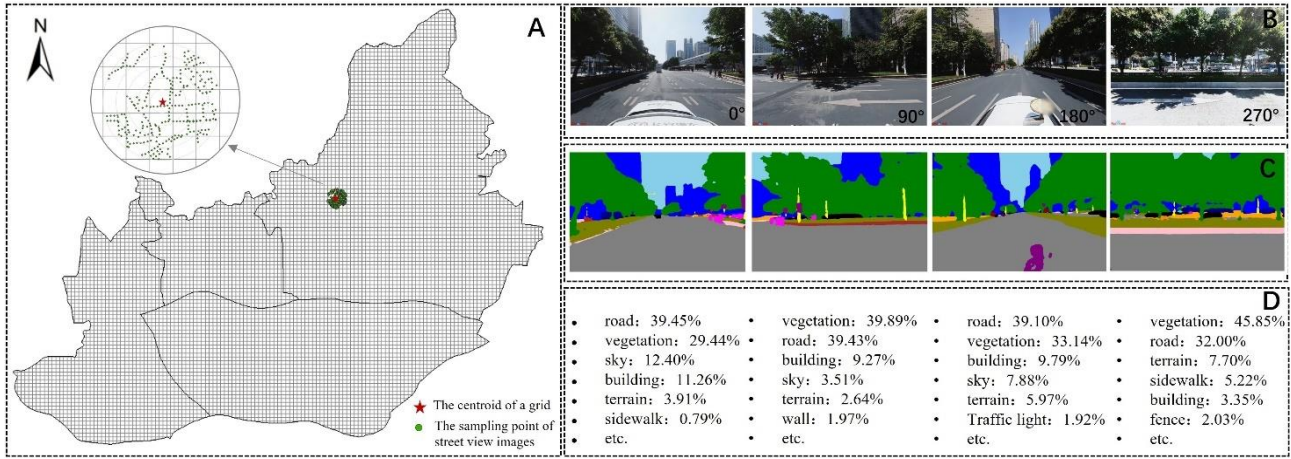

Fig.1 Illustration of street view sampling and feature extraction process. (A) The process of sampling street view images around the centroid of a randomly selected grid in a circular buffer with radii ranging from 100m to 500m. (B) The retrieval of sampling images in four different directions (0°, 90°, 180°, 270°) at a random sampling location. (C) Feature extraction utilizing the Mask2Former scene parsing algorithm. (D) The proportion of features across 19 segmented classes in each image.

For feature extraction, we used Mask2Former, a state-of-the-art deep learning method that outperforms various datasets. Based on a Transformer architecture with self-attention mechanisms, it captures long-range dependencies and global information more effectively, especially in complex scenes and multiple segmentation tasks (Cheng et al., 2022). The algorithm assigned each pixel in an image with pixel-wise semantic labels across 19 classes: terrain, vegetation, sky, wall, building, road, traffic sign, traffic light, sidewalk, fence, pole, bus, train, truck, car, bicycle, motorcycle, rider, and person. These were categorized into four broad groups: 1) transport vehicles (bus, car, truck, motorcycle); 2) transport network (road, sidewalk, traffic light, traffic sign, pole, on rails); 3) human activity (building, wall, fence, rider, bicycle, person); 4) natural environment (vegetation, sky, terrain). The percentage of each feature in an SVI was estimated, and the mean ratio for each feature among SVIs in a specific buffer represented the general characteristics at each aggregation location, serving as on-road predictor variables for empirical air quality models.

*2.3 Prediction and Evaluation*

Three typical machine learning methods: XGBoost (Chen and Guestrin, 2016), random forest (Qi et al., 2022), and neural network were employed to perform the same prediction tasks under consistent conditions. Also, the step-wise linear regression method was carried out for direct comparison with machine learning methods (Hankey et al., 2019). Four regulatory pollutants, including NO, $NO_2$, $PM_{2.5}$, and $PM_{10}$, were forecasted to assess the effects of algorithms, sampling strategy and SVI quality on prediction errors. Thus, a standardized prediction pipeline with consistent model parameters and evaluation metrics was necessary.

We conducted a grid search to optimize the critical hyperparameters for each model. For XGBoost, we searched for the optimal values of eta (0, 0.01, 0.05, 0.1, 0.3, 0.5), minimum child weight (1, 2, 3, 4, 5), maximum tree depth (3, 5, 7, 9, 10), gamma (0, 0.1, 0.3, 0.5, 0.7, 0.9, 1), and subsample (0.5, 0.6, 0.7, 0.8, 0.9, 1). Similarly, for the random forest algorithm, we searched for the optimal number of decision trees (50, 100, 200, 500) and maximum tree depth (6, 7, 8, 9, 10, 11, 12). For neural networks, we examined the optimal learning rate (0.0001, 0.0003, 0.0005, 0.001) and batch size (8, 16, 32, 64). These thorough processes ensure the result accuracy.

Five widely used evaluation metrics were introduced to qualify prediction errors, including square error (MSE), mean absolute error (MAE), root mean squared error (RMSE), mean absolute percentage error (MAPE), and cross-validation (CV) $R^2$ as metrics. This choice of evaluation metrics, widely accepted in the field, adds to the validity of our study. To assess model robustness and generalizability, we conducted 10-fold CV for every model by randomly dividing the original dataset into 10 subsets. Each subset was held out in turn, while the remaining nine subsets were used to construct a model and predict the data for the held-out subset. This process was repeated 10 times, and the mean $R^2$ was reported as the final 10-fold CV $R^2$.

**3. Results and discussion**

*3.1. Descriptive Statistics*

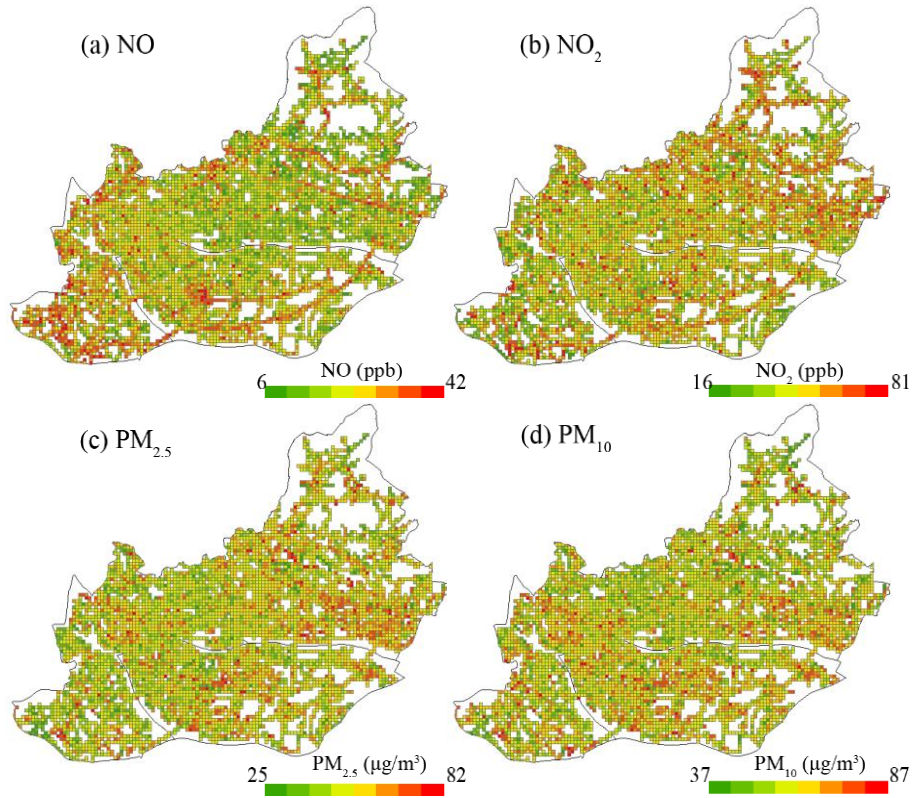

Fig.2 Spatial distribution of four pollutants concentrations

Figure 2 illustrates the spatial distribution of concentrations for four pollutants within a 200-mr resolution. Notably, high concentrations of NO and $NO_2$ exhibit a "line-like" distribution, primarily located along major roads at the edge of the central urban area, like roads in the southern Liwan District, primary roads in the middle of Haizhu District, and the expressway in Tianhe District. In contrast, high concentrations of $PM_{2.5}$ and $PM_{10}$ show a "block-like" distribution, mainly concentrated in commercial areas and densely populated old urban districts and villages. This pattern is consistent with population distribution patterns identified by (Zhong et al., 2023), indicating that population activities likely influence PM concentrations.

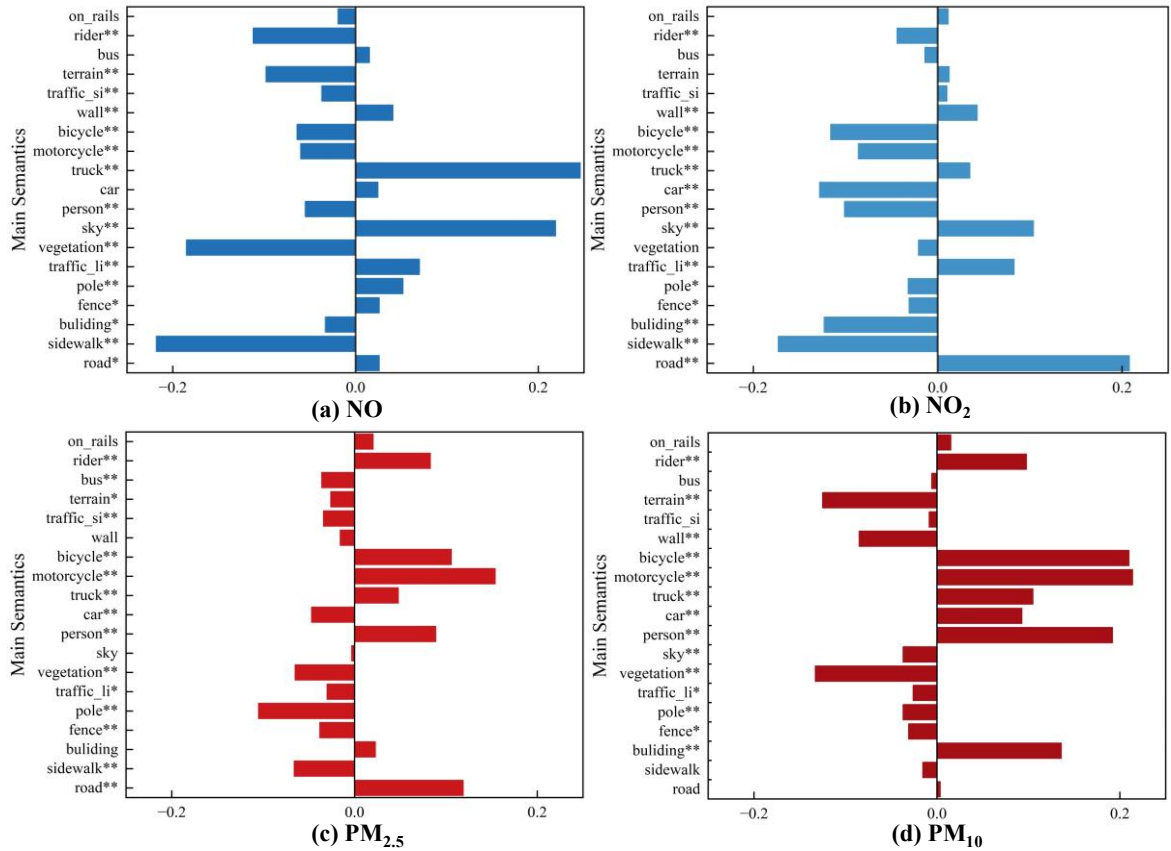

Fig.3 The correlation between the air pollutant concentrations and SVI features

Fig. 3 presents the correlations between the four pollutants and the 19 categories of SVI feature. A strong correlation exists between NO concentrations and truck features, with a similar spatial distribution (Fig. S1), while peak correlation for $NO_2$ appears at 'road' feature. This is suggested that traffic-related factors, particularly trucks, significantly contribute to $NO_x$ pollution, aligning with findings by Harrison et al. (2021). Additionally, NO and NO2 show a consistent trend with 'sky' features and an opposite trend with 'building' features. This finding supports the spatial distribution of NO and NO2, where higher ratios of natural features correlate with lower human density, while more significant 'building' features denote more frequent human activity (Nathvani et al., 2023). Comparably, PM concentrations correlate positively with traffic-related features (motorcycles and road) and human activity features (rider, person, and building), while they correlate negatively with natural environment features (sky and vegetation). This result implies that high PM concentrations are attributed to higher human density since a higher proportion of sky typically indicates broad spaces with fewer obstructions from buildings.

*3.2 The Effects of Different Algorithms*

In the model development stage, we compared the performance of three typical machine learning algorithms using a consistent image collection and processing protocol, as shown in Table 1. To ensure a fair comparison, we selected SVIs sampled at 0° within a 200-m radius buffer. The same datasets with consistent conditions were applied across different experiments. Besides, the predictor variables used in the previous LUR model were identical to those of other models (Hankey and Marshall, 2015).

Overall, the machine learning methods generally outperformed LUR, ranking as follows: random forest > XGBoost > neural network > LUR. For all pollutants, random forest consistently outperformed the others, demonstrating lower error metrics (MSE, MAE, RMSE, MAPE) and higher goodness-of-fit measures (10-fold CV $R^2$). XGBoost and neural networks showed comparable performance, with the former having slightly lower errors. When evaluated using MSE as the metric, XGBoost, random forest, and neural networks outperformed the simple multiple linear regression model, with average improvements of 3.93%, 6.13%, and 3.22%, respectively. Notably, these algorithms showed prominent improvements in the NO prediction task, upgrading performance by 7.62%, 11.28%, and 5.97%, respectively. The relatively low $R^2$ results suggest there is not a simple linear relationship between on-road air pollution and SVI features. Nonlinear models, such as the selected machine learning algorithms, are more suitable for predicting pollutant concentrations. Whereas, this SVI-based LUR model considered more detailed street-level features compared to traditional GIS-based LUR models, which always tend to underestimate pollution in polluted areas and overestimate it in clean places due to omission of local predictor variables (Cordioli et al., 2017; Larkin et al., 2017).

Table 1 Prediction performance of different models

| Pollutants | Algorithm | MSE | MAE | RMSE | MAPE | 10-fold CV $R^2$ |
|---|---|---|---|---|---|---|
| NO | XGBoost | 9.443 | 2.319 | 3.070 | 8.325 | 0.198 |
| | Random Forest | **9.132** | **2.260** | **3.019** | **8.240** | **0.225** |
| | Neural Network | 9.590 | 2.351 | 3.096 | 8.892 | 0.186 |
| | Previous LUR | 10.162 | 2.431 | 3.185 | 8.412 | 0.137 |
| $NO_2$ | XGBoost | 18.189 | 3.213 | 4.263 | 4.910 | 0.103 |
| | Random Forest | **17.854** | **3.180** | **4.224** | **4.764** | **0.106** |
|  | Neural Network | 18.280 | 3.225 | 4.274 | 4.854 | 0.098 |
|  | Previous LUR | 18.467 | 3.244 | 4.296 | 4.909 | 0.101 |
|  | XGBoost | <u>8.422</u> | <u>2.220</u> | <u>2.901</u> | <u>3.602</u> | <u>0.076</u> |
| PM$_{2.5}$ | Random Forest | **8.267** | **2.193** | **2.874** | **3.526** | **0.093** |
|  | Neural Network | 8.507 | 2.237 | 2.916 | 3.636 | 0.069 |
|  | Previous LUR | 8.825 | 2.288 | 2.970 | 3.710 | 0.032 |
|  | XGBoost | <u>8.950</u> | <u>2.298</u> | <u>2.991</u> | <u>2.979</u> | <u>0.076</u> |
| PM$_{10}$ | Random Forest | **8.887** | **2.289** | **2.980** | **2.943** | **0.082** |
|  | Neural Network | 9.008 | 2.311 | 3.001 | 3.031 | 0.072 |
|  | Previous LUR | 9.158 | 2.336 | 3.025 | 2.990 | 0.055 |

Note*: The best performance in each pollutant prediction task is highlighted in boldface, and the second-best performance is underlined.

## 3.3. Sampling Strategy for SVIs

### 3.2.1 Sampling Angle

To explore the effects of sampling angles for SVIs on prediction errors, we obtained images from 0°, 90°, 180°, and 270° at the same locations. All experiments used identical datasets with SVIs sampled within a 200-meter radius buffer to ensure a fair comparison. Fig. 4 shows the prediction errors of four algorithms with SVIs sampled at different angles under the same radius buffer.

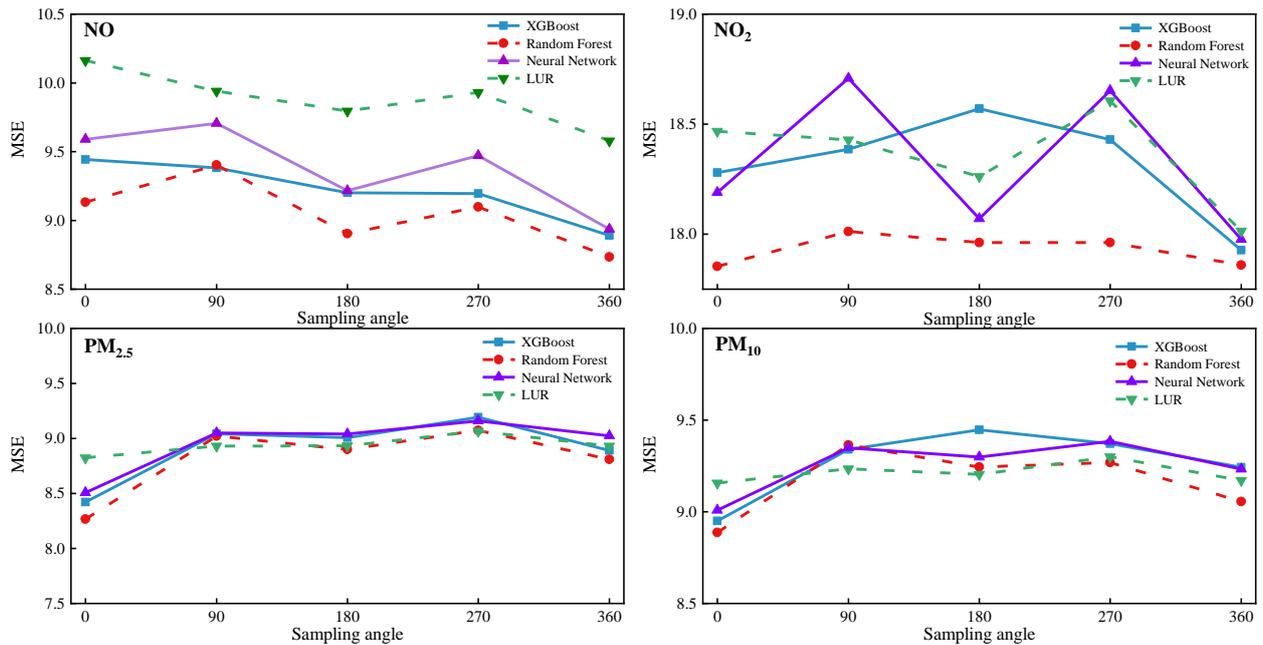

Fig.4 Impact of sampling angle on prediction performance. Note: 360° means the averaging strategy of SVIs features.

Generally, for on-road pollution prediction, SVI-based models using images sampled at 0° and 180°angles presented consistently lower prediction errors than others, with performance improvements of up to 9.1%. Despite similar SVI features at the same location (Qi et al., 2022), significant content variations exist in images from different sampling angles. Typically, 0° and 180° SVIs capture views in the forward and reverse directions along the road, naturally producing traffic-related features (e.g., trucks, cars, roads, sidewalks). Conversely, 90° and 270° SVIs record the surrounding environment, often extracting more features related to human activity (e.g., buildings, walls, fences) (Fig.5). However, regardless of the sampling angle, the top five features were road, vegetation, building, sky, and car, encompassing traffic networks, vehicles, building environments, and natural environments (Qi et al., 2022; Zhang et al., 2023).

Some studies proposed averaging strategy for SVI features from different angles to diminish bias caused by single-angle SVIs (Qi and Hankey, 2021; Xu et al., 2022). This method averages the SVI feature ratios from 0°, 90°, 180°, and 270° to reduce feature. In Fig. 4, 360° means this prediction performance using averaging strategy. For NO and $NO_2$, this approach showed superior performance, with all models achieving optimal results. Random forest remained the best-performing model, with MAEs of 2.23 and 3.14, respectively. However, for $PM_{2.5}$ and $PM_{10}$, the single-angle strategy with 0° yielded the best results. These findings suggest that while the averaging strategy for SVI features significantly reduces errors resulting from individual angles, its effectiveness is task-dependent.

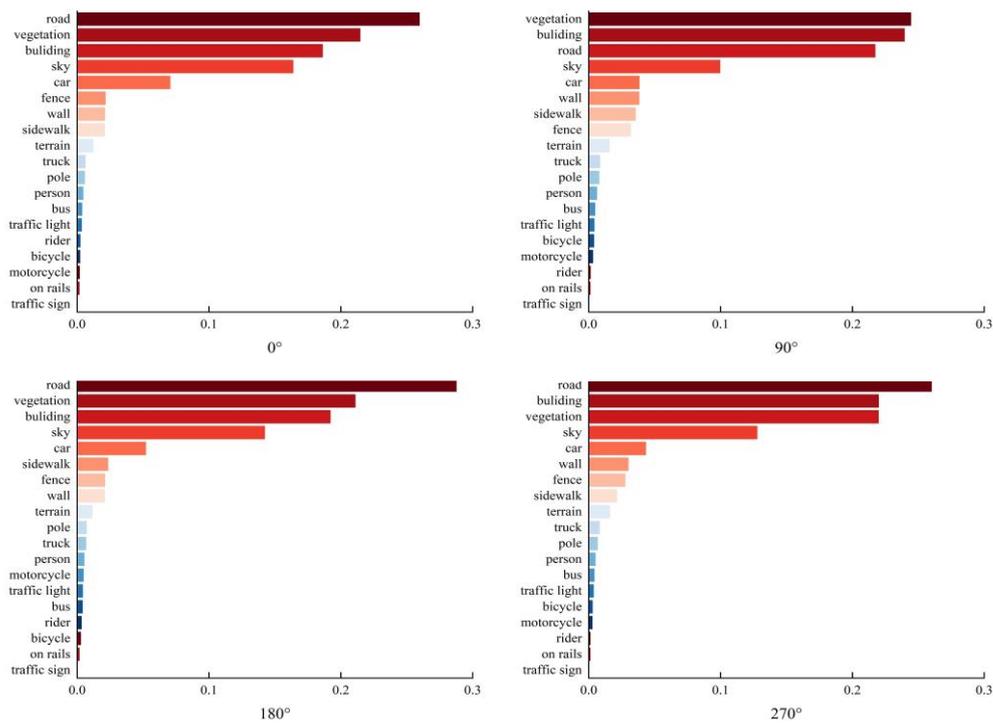

Fig.5 Ranking of segmented feature ratios at different sampling angles

*3.2.2 Sampling Range*

Representing local characteristics related to on-road pollution using appropriate SVI features is crucial. An excessive number of features from too large or too small a range can introduce bias, leading to increased prediction errors. To explore the effects of sampling range, we used the averaging strategy for SVIs sampled within buffer sizes ranging from 100 to 500 m at 100 m intervals. This resulted in five different sub-datasets of SVI features due to the varying circular buffer sizes. All data processing and procedures remained consistent across experiments. Figure 6 depicts the model performance across various sampling ranges, while Fig.S2 shows comparative results using 0° SVIs for the same prediction tasks. The results underscore the importance of the averaging-strategy SVIs, which slightly outperformed the 0° SVIs. This not only validates the feasibility of the averaging strategy but also suggests its potential to enhance model performance.

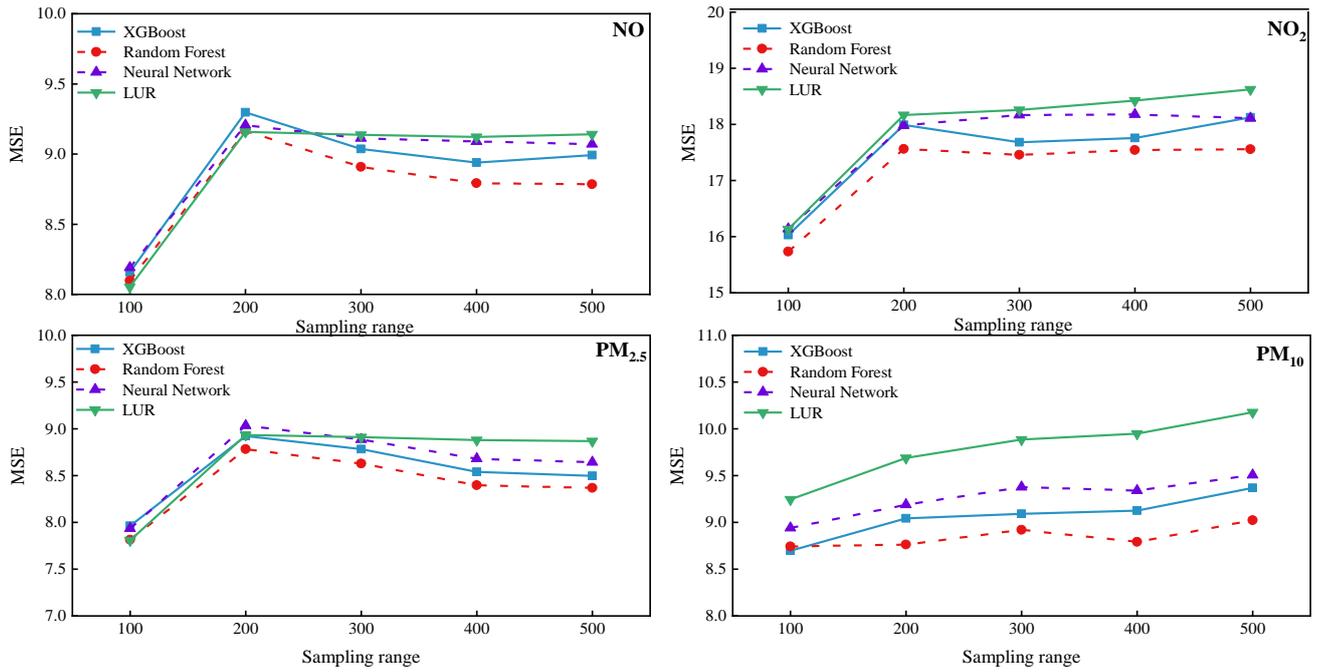

Fig.6 Impact of sampling range on prediction performance

It is worth noting that estimation errors increased as the sampling buffer size expanded. A particularly dramatic error surge was observed when the radius was increased from 100 to 200 m, resulting in the worst prediction outcomes for NO and $PM_{2.5}$. For $NO_2$ and $PM_{10}$ predictions, the bias continued to fluctuate upwards with the increasing radius. Introducing more irrelevant features with the extension of the buffer size led to a reduction in prediction accuracy (Hong et al., 2019; Qi and Hankey, 2021). Random forest consistently emerged as the best method across all tasks, with improved MSEs of 10.1%, 15.4%, 13.6%, and 13.5% over other models in predicting NO, $NO_2$, $PM_{2.5}$, and $PM_{10}$ concentrations, respectively.

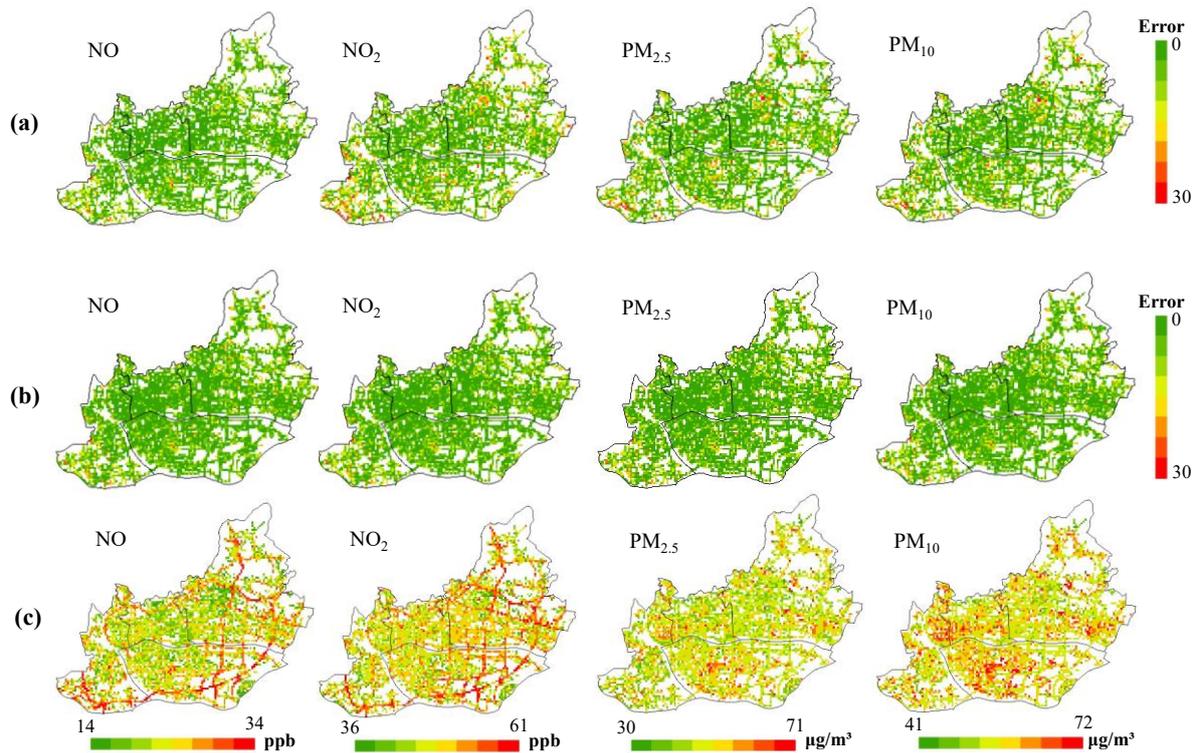

Fig.7 Distribution of errors between prediction and actual concentrations using (a) the mean SVI features sampled by averaging strategy; (b) the mean SVI features sampled by averaging strategy in a buffer with radii 100m; (c) Spatial distribution of prediction results using random forest.

To provide a more comprehensive qualitative analysis, we calculated the absolute prediction errors between the predicted and true values using the random forest algorithm (Fig. 7). Prediction errors significantly decreased when using the optimal strategy for angle and range. Model performance improved by 4.44%, 13.48%, 5.80%, and 9.74% for the four pollutants, respectively, using SVI features sampled within a 100-m radius buffer with the averaging strategy. Furthermore, combined with the prediction results (Figure 7(c)), it is evident that the model generally tends to overestimate pollution concentrations, especially at the edge of urban areas. However, the overall prediction bias is at most 2.5 μg/m³ or ppb, except for local high-pollution places. This ability of the model to adapt to different pollution levels not only demonstrates its robustness but also suggests its potential for real-world applications (Cai et al., 2020; Ge et al., 2022).

*3.4 Low-quality SVIs*

Using the image quality recognition algorithm, we identified 57,790 low-quality images, accounting for 16.25% of the total SVIs. The four common low-quality conditions and their proportions are blur (1.47%), underexposure (0.29%), overexposure (47.21%), and color channel distortion (51.04%). Fig.S4 presents the frequency distribution of SVIs and low-quality SVIs at aggregation locations. Generally, each aggregation location contains at least 20 SVIs, with a low-quality SVI proportion ranging from 10% to 30%. To qualitatively analyze the effects of image quality, we selected data with low-quality SVIs at corresponding aggregation locations. The dataset was divided into six groups based on the proportion of low-quality SVIs at each aggregation location: 0%, 0-5%, 5%- 10%, 10%- 15%, 15%- 20%, and 20%- 30%.

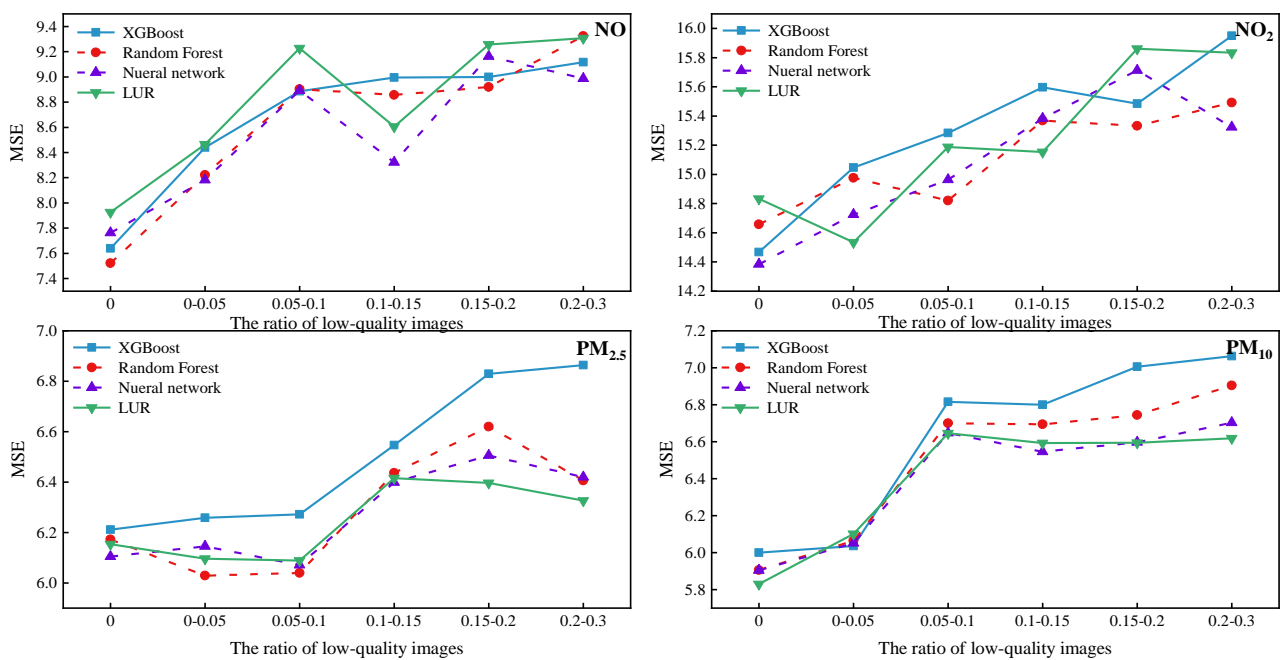

Fig.8 Impact of low-quality SVIs on prediction performance

Fig.8 demonstrates the prediction performance for different ratios of low-quality SVIs. Overall, as the proportion of low-quality SVIs increases, prediction errors show a wavy rising trend. Filtering out all low-quality SVIs significantly improved prediction accuracy by 23.95%, 10.08%, 23.44%, and 21.15% for NO, $NO_2$, $PM_{2.5}$, and $PM_{10}$, respectively, compared to the worst condition. Fig.9 displays the feature extraction results for the four low-quality SVI conditions. Visually, overexposure, blur, and underexposure conditions render the images nearly unrecognizable, leading to misjudgments and incorrect identifications by the deep learning method. For example, the proportion of 'road' features drastically increases to more than twice the normal level, while features related to human activities, such as buildings, sky, and cars, are erased and unrecognized. In contrast, SVIs suffering from color channel distortion still maintain relatively high accuracy during feature extraction. Table S1 summarizes the estimation errors for low-quality SVIs with four issues. Due to varying dataset sizes for each condition, there is some bias in the comparison results. However, the model performance under different conditions ranks as follows: color channel distortion > blur and underexposure > overexposure. Notably, the previous LUR model almost fails in all prediction tasks, with the worst performance.

Based on the above analysis, the optimal sampling strategy combines the sampling angle (averaging strategy) and range (100m radius buffer). Fig.S6 contrastively demonstrates prediction errors using the optimal sampling strategy, whether filtering low-quality SVIs or not. The results show improved performance across all models after excluding aggregation locations with a low-quality SVI proportion of more than 20%. The best-performing models demonstrated improvements of 7.35%, 9.44%, 7.98%, and 8.17% for NO, $NO_2$, $PM_{2.5}$, and $PM_{10}$, respectively, compared to the worst-performing models. These findings underscore the importance of street view image quality and sampling strategies for model performance. Qi et al. (2022) observed that mobile monitoring data coupled with SVI provides a potential for estimating local air pollution, which traditional GIS-based methods cannot capture (Xu et al., 2021). With the growing availability of public streetscape imagery, leveraging these resources to build large-scale empirical air quality models (e.g., national or global) under a unified framework holds significant promise (Hu et al., 2020; Biljecki and Ito, 2021; Sabedotti et al., 2023). These findings provide valuable references and support in developing image-based air pollution model and SVI-based studies.

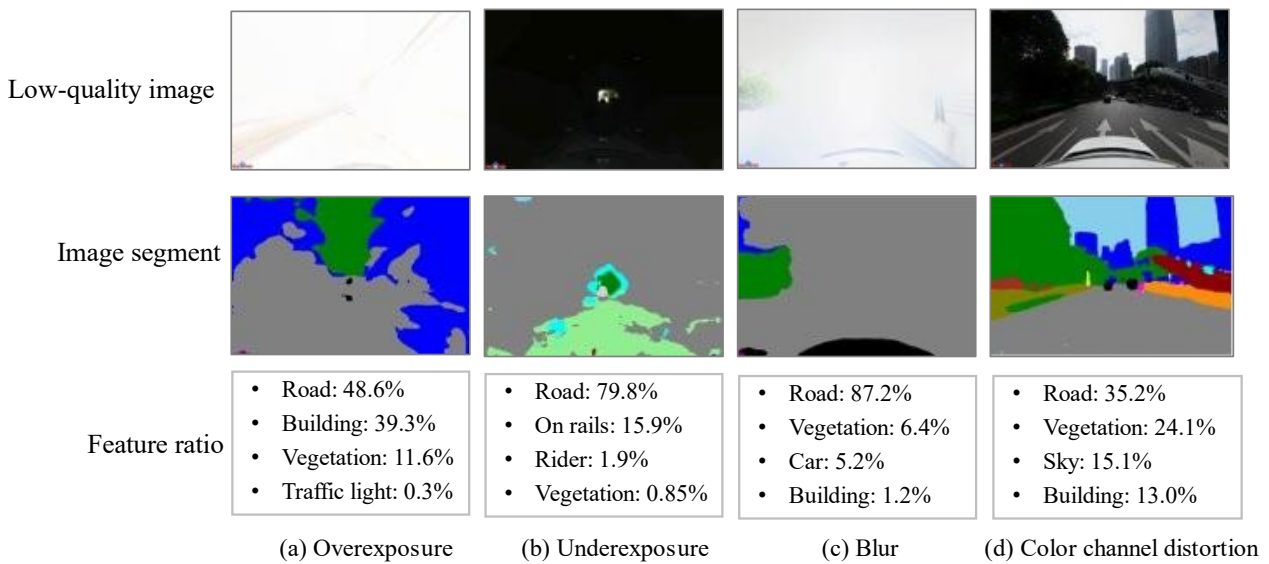

Fig.9 Segmented features in low-quality images under overexposure, underexposure, blur and color channel distortion.

## 4. Conclusion

This study utilized 314 taxis to dynamically monitor on-road pollution and sample corresponding SVIs, aiming to quantify the biases and errors introduced by algorithms, sampling strategies, and image quality in SVI-based models for predicting on-road air pollution. We extracted SVI features from approximately 382,000 streetscape images, considering sampling angles (0°, 90°, 180°, 270°) and buffer ranges (radii of 100m, 200m, 300m, 400m, 500m). Three machine learning algorithms (random forest, XGBoost, neural network) and the linear LUR model were evaluated for their prediction performance. Additionally, we identified and analyzed the impact of four typical image quality issues: blur, underexposure, overexposure, and color channel distortion.

The results reveal that high concentrations of NO and $NO_2$ exhibit a "line-like" distribution at the edges of the study area, attributed to traffic-related features (e.g., trucks, roads, traffic lights). Conversely, high concentrations of $PM_{2.5}$ and $PM_{10}$ display a "block-like" distribution influenced by human activity and traffic-related features. Quantitatively, machine learning methods generally outperformed the linear LUR model, with the ranking as follows: random forest > XGBoost > neural network > LUR. Compared to traditional GIS-based LUR models that often produce underestimation or overestimation, the SVI-based LUR model incorporates more detailed street-level features. For single-angle sampling strategies, models using 0°-SVIs outperformed others, while the averaging strategy is more effective avoiding bias brought by limited single angle. Furthermore, estimation errors increased with larger buffer sizes due to introducing more irrelevant features. Therefore, model using SVIs obtained within a 100m radius buffer outperforms than others. Naturally, the optimal sampling strategy is to obtain SVIs at a 100m radius buffer and extract features using the averaging strategy. This achieved estimation results for each aggregation location with absolute errors less than 2.5 μg/m³ or ppb. As the proportion of low-quality SVIs increased in aggregation locations, prediction errors showed a wavy rising trend. Issues such as overexposure, blur, and underexposure led to misjudgments and incorrect identifications by the deep learning method, causing an overrepresentation of road features and unrecognized features related to human activity. Filtering out all low-quality SVIs significantly improved prediction accuracy by 23.95%, 10.08%, 23.44%, and 21.15% for NO, $NO_2$, $PM_{2.5}$, and $PM_{10}$, respectively, compared to the worst condition. With the growing availability of public streetscape imagery, leveraging these resources to build large-scale empirical air quality models holds significant promise. These findings provide a better understanding and valuable support for developing image-based air quality models and other SVI-related research.

**Acknowledgements**

This study was funded by the Nansha District Key Research and Development Project (No.2023ZD006); and Guangzhou Municipal Ecological Environment Bureau (No. K22-76160-026).

**Appendix A. Supplementary material**

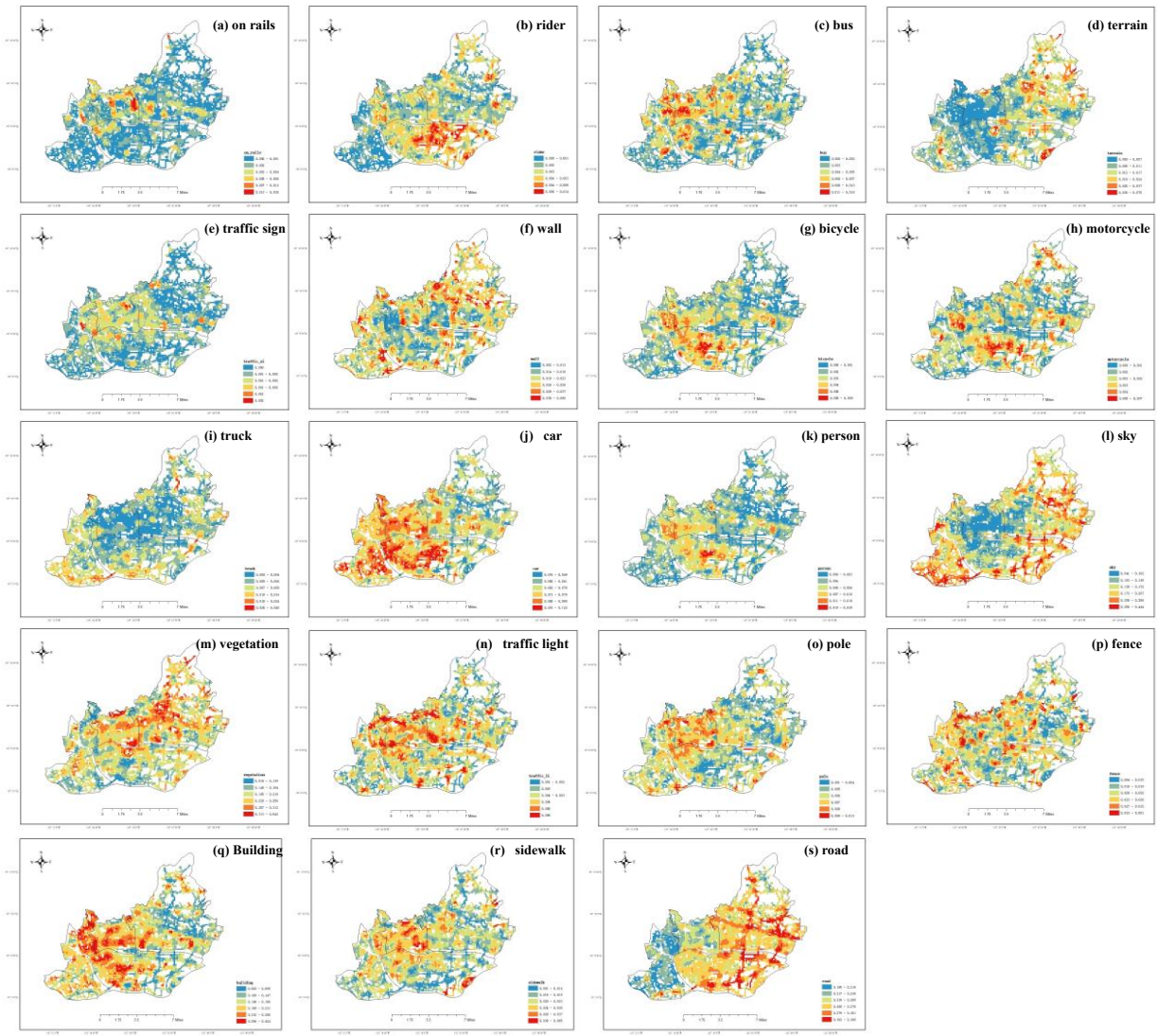

Fig.S1 Spatial distribution of the mean ratio of 19 segmented categories

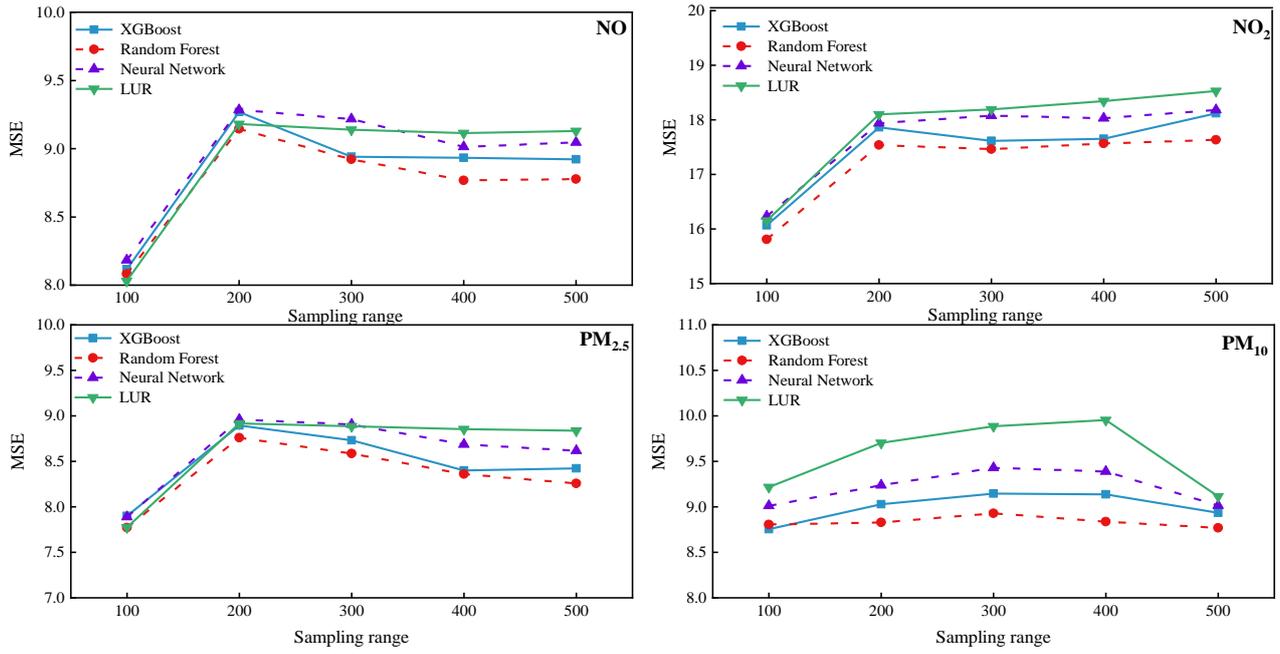

Fig.S2 Impact of sampling range on prediction performance using 0°-SVIs

Table S1 Prediction performance of low-quality images under overexpose, underexposure, blur and color channel distortion.

| Pollutants | Algorithm | Blurry and underexposure | | | Overexposed | | | Color channel distortion | | |
|---|---|---|---|---|---|---|---|---|---|---|
| | | MSE | MAE | RMSE | MSE | MAE | RMSE | MSE | MAE | RMSE |
| NO | XGBoost | 9.85 | 2.33 | 3.03 | 14.59 | 2.88 | 3.66 | 7.99 | 2.13 | 2.78 |
| | Random Forest | 10.18 | 2.39 | 3.11 | 14.19 | 2.74 | 3.60 | 8.38 | 2.22 | 2.85 |
| | Neural Network | **9.14** | **2.44** | **3.02** | **13.92** | **2.92** | **3.73** | **8.61** | **2.32** | **2.93** |
| | LUR | 40.77 | 3.07 | 4.80 | 19.66 | 3.14 | 4.18 | 12.19 | 2.65 | 3.42 |
| $NO_2$ | XGBoost | 25.04 | 3.52 | 4.80 | 42.50 | 4.56 | 6.37 | 18.41 | 3.10 | 4.13 |
| | Random Forest | 23.72 | 3.25 | 4.64 | 43.93 | 4.66 | 6.48 | 19.53 | 3.18 | 4.21 |
| | Neural Network | **20.53** | **3.45** | **4.53** | **41.74** | **4.71** | **6.46** | **17.54** | **3.23** | **4.19** |
| | LUR | 36.48 | 3.76 | 5.68 | 43.59 | 4.81 | 6.51 | 27.00 | 3.69 | 4.83 |
| $PM_{2.5}$ | XGBoost | 18.53 | 2.75 | 3.98 | 85.62 | 5.24 | 8.24 | 12.87 | 2.45 | 3.26 |
| | Random Forest | 18.06 | 2.67 | 3.92 | 78.86 | 5.15 | 7.98 | 12.50 | 2.40 | 3.23 |
| | Neural Network | **17.12** | **2.98** | **4.13** | **75.82** | **6.31** | **8.69** | **11.14** | **2.62** | **3.33** |
| | LUR | 28.23 | 3.22 | 4.87 | 103.52 | 6.33 | 9.41 | 13.22 | 2.55 | 3.35 |

|  | | | | | | | | | |
|---|---|---|---|---|---|---|---|---|---|
|  | XGBoost | 21.22 | 2.85 | 4.23 | 70.04 | 5.06 | 7.56 | 10.65 | 2.52 | 3.13 |
| PM$_{10}$ | Random Forest | 21.33 | 2.90 | 4.26 | 61.32 | 4.70 | 7.10 | 10.66 | 2.49 | 3.13 |
|  | Neural Network | **19.31** | **3.23** | **4.38** | **60.97** | **5.55** | **7.80** | **9.58** | **2.59** | **3.09** |
|  | LUR | 29.05 | 3.44 | 5.08 | 84.08 | 5.91 | 8.63 | 11.96 | 2.69 | 3.33 |

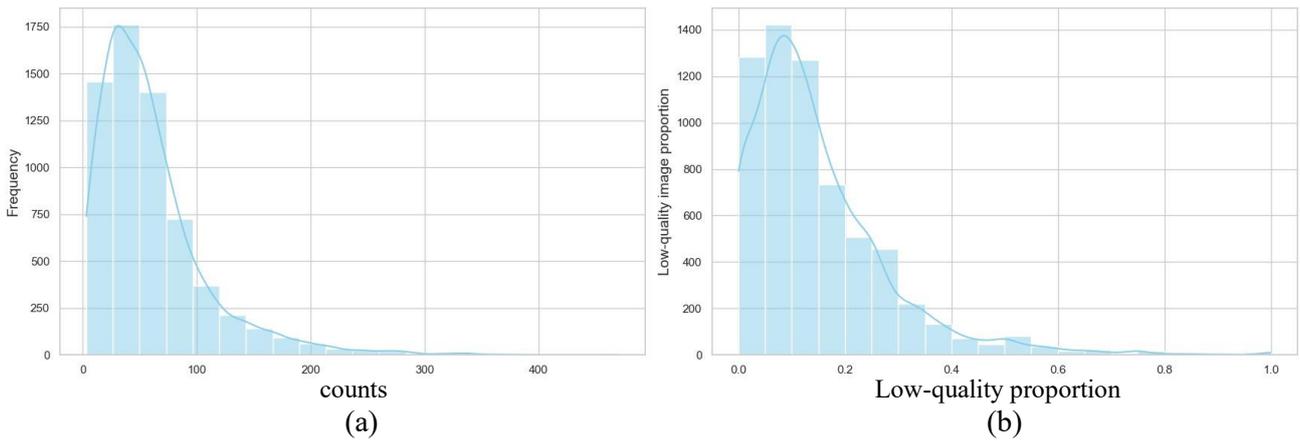

Fig.S3 Frequency distribution of (a) SVIs counts and (b) low-quality SVI proportion

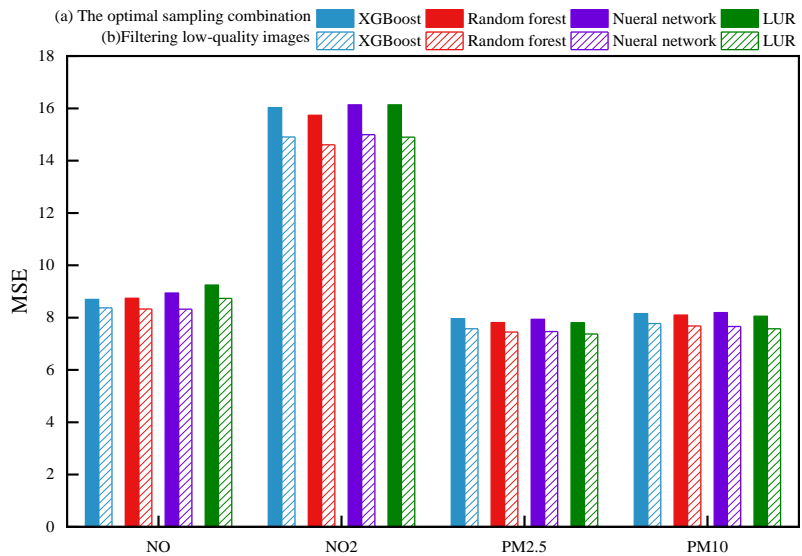

Fig.S4 Prediction Performance whether considering image quality. (a) Using the optimal sampling combination (SVIs sampled at a 100m radius buffers with averaging strategy); (b) Filtering aggregation locations with proportion of low-quality SVIs more than 20% based on the optimal sampling combination.

Google Street View-Derived Urban Greenspace and Google Air View-Derived Pollution Levels, Environ Sci Technol, 57, https://doi.org/10.1016/acs.est.3c05000